\documentclass{article}
\usepackage{ijcai26}

\usepackage{times}
\usepackage{soul}
\usepackage{url}
\usepackage[hidelinks]{hyperref}
\usepackage[utf8]{inputenc}
\usepackage[small]{caption}
\usepackage{graphicx}
\usepackage{amsmath}
\usepackage{amsthm}
\usepackage{amssymb}
\usepackage{booktabs}
\usepackage{algorithm}
\usepackage{algorithmic}
\usepackage[switch]{lineno}
\usepackage{multirow}
\usepackage{color}
\usepackage{pifont}
\usepackage[dvipsnames,svgnames,x11names]{xcolor}
\usepackage{subfig}
\definecolor{sridyellow}{RGB}{255, 250, 140}  % 对应图中的淡黄色
\definecolor{sridgreen}{RGB}{205, 240, 145}   % 对应图中的嫩绿色
\definecolor{sridpurple}{RGB}{200, 160, 255}  % 对应图中的淡紫色
\title{A Survey on 3D Skeleton Based Person Re-Identification: Taxonomy, Advances, Challenges, and Interdisciplinary Prospects}

\author{
    Author Name
    \affiliations
    Affiliation
    \emails
    email@example.com
}

\author{
Haocong Rao$^{1,2,3}$\And
Chunyan Miao$^{1,2,3}$\footnotemark[1]
\affiliations
$^{1}$ College of Computing and Data Science, Nanyang Technological University (NTU), Singapore\\
$^{2}$ Joint NTU-UBC Research Centre of Excellence in Active Living for the Elderly (LILY), NTU, Singapore\\
$^{3}$ Alibaba-NTU Global e-Sustainability CorpLab (ANGEL), NTU, Singapore
\emails
\{haocong.rao, ascymiao\}@ntu.edu.sg
}
\begin{document}

\maketitle
\footnotetext[1]{Corresponding Author}
\begin{abstract}
Person re-identification via \textit{3D skeletons} is an important emerging research area that attracts increasing attention within the pattern recognition community. With distinctive advantages across various application scenarios, numerous 3D skeleton based person re-identification (SRID) methods with diverse skeleton modeling and learning paradigms have been proposed in recent years.  
In this paper, we provide a comprehensive review and analysis of recent SRID advances. First of all, we define the SRID task and provide an overview of its origin and major advancements.
Secondly, we formulate a systematic taxonomy that organizes existing methods into three categories centered on hand-crafted, sequence-based, and graph-based modeling. Then, we elaborate on the representative models along these three types with an illustration of foundational mechanisms. Meanwhile, we provide an overview of mainstream supervised, self-supervised, and unsupervised SRID learning paradigms and corresponding common methods. A thorough evaluation of state-of-the-art SRID methods is further conducted over various types of benchmarks and protocols to compare their effectiveness, efficiency, and key properties.   
Finally, we present the key challenges and prospects to advance future research, and highlight interdisciplinary applications of SRID with a case study.
A curated collection of valuable resources is available at \href{https://github.com/Kali-Hac/3D-SRID-Survey}{https://github.com/Kali-Hac/3D-SRID-Survey}.

\end{abstract}

% \cite{vezzani2013people,zheng2015towards,nambiar2019gait,wu2020rgb,ye2021deep,haque2016recurrent,karianakis2018reinforced,cho2022part}.
\section{Introduction}
Person re-identification (re-ID) is an essential pattern recognition task of matching and retrieving a person-of-interest across different views or scenes, which has been widely applied to security authentication, smart surveillance, activity monitoring, healthcare, and embodied AI \cite{nambiar2019gait,ye2021deep,bao2026activityforensics}. Recent economical and precise skeleton-tracking devices ($e.g.,$ Kinect \cite{shotton2011real-time}) have simplified the acquisition of 3D skeleton data, enabling them to be a prevalent and versatile data modality for gait analysis and person re-ID \cite{liao2020model,rao2024hierarchical}. Unlike conventional person re-ID methods that rely on appearance or facial characteristics \cite{ye2021deep},
% \cite{wang2016person,zhao2017person,su2018multi,li2019unsupervised,zhang2019densely,lan2020semi,ge2020selfpaced,cho2022part}, 
3D \textbf{S}keleton based person \textbf{R}e-\textbf{ID} (SRID) models typically exploit body-structure features and motion patterns ($e.g.,$ gait \cite{murray1964walking}) from 3D positions of key body joints to identify different persons. 
With unique merits such as small input data, light-weight models, privacy-preserving without using appearances, and robustness against view and background variations \cite{han2017space}, SRID has attracted surging attention from both academia and industry \cite{rao2021self}.

\begin{figure}
    \centering
\includegraphics[width=0.99\linewidth]{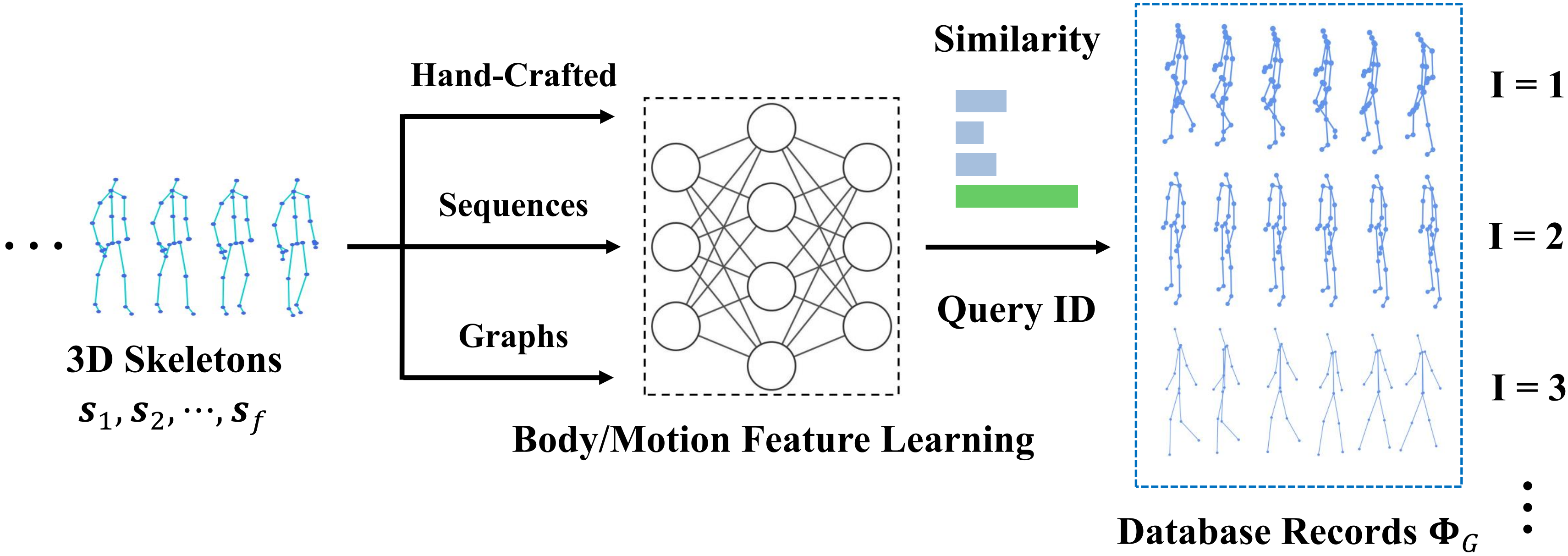}
    \caption{Overview of 3D skeleton based person re-ID (SRID) task with hand-crafted, sequence-based or graph-based modeling to learn effective body and motion features for identity recognition.}
    \label{task_overview}
\end{figure}

\begin{figure*}[t]
    % 0.75
      \subfloat[Origin and Advancements of SRID Research]{
 	 \begin{minipage}[b]{0.64\textwidth}
       \centering
       \scalebox{0.44}{
 	   \includegraphics[]{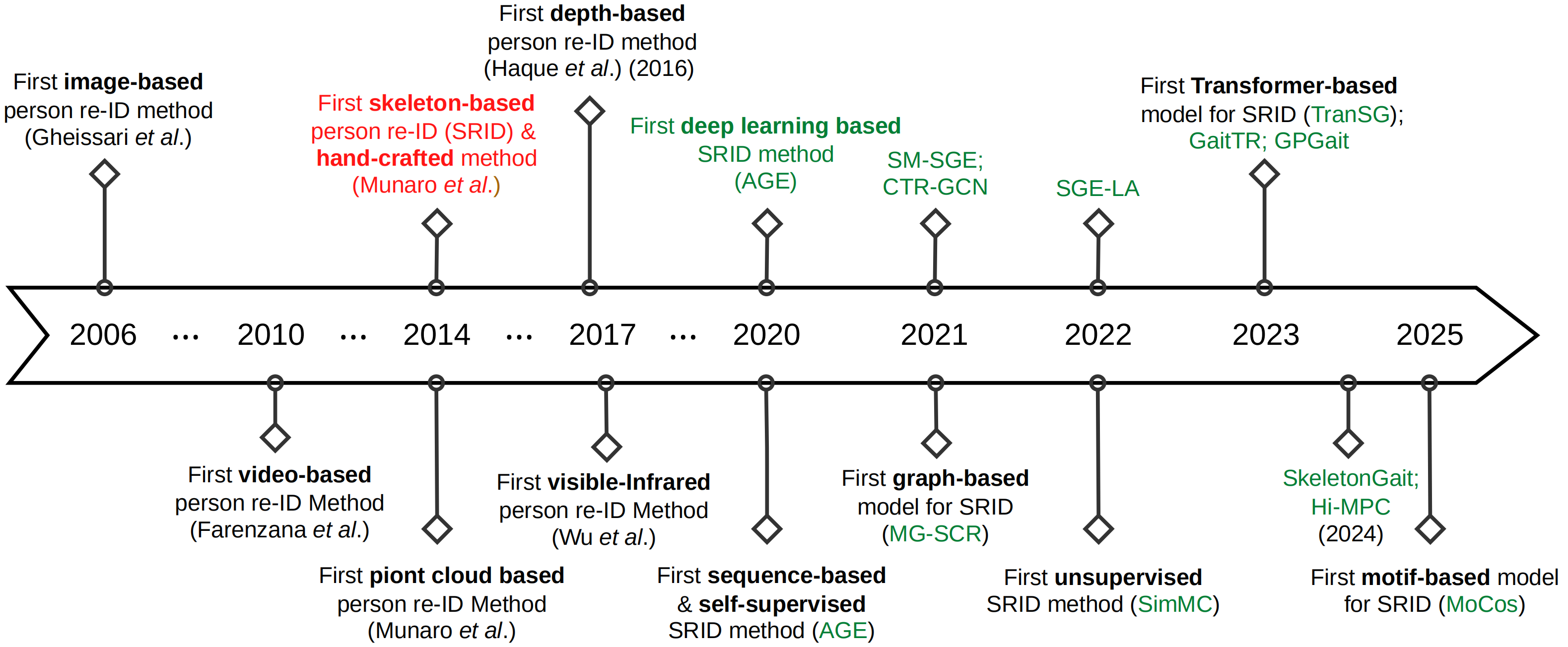}
     }
 	  \end{minipage}
 	\label{SRID-Timeline}
 	}
  % 0.45
  \
   \subfloat[Model Accuracy \& Efficiency Comparison]{
 	 \begin{minipage}[b]{0.35\textwidth}
       \centering
       \scalebox{0.155}{
 	   \includegraphics[]{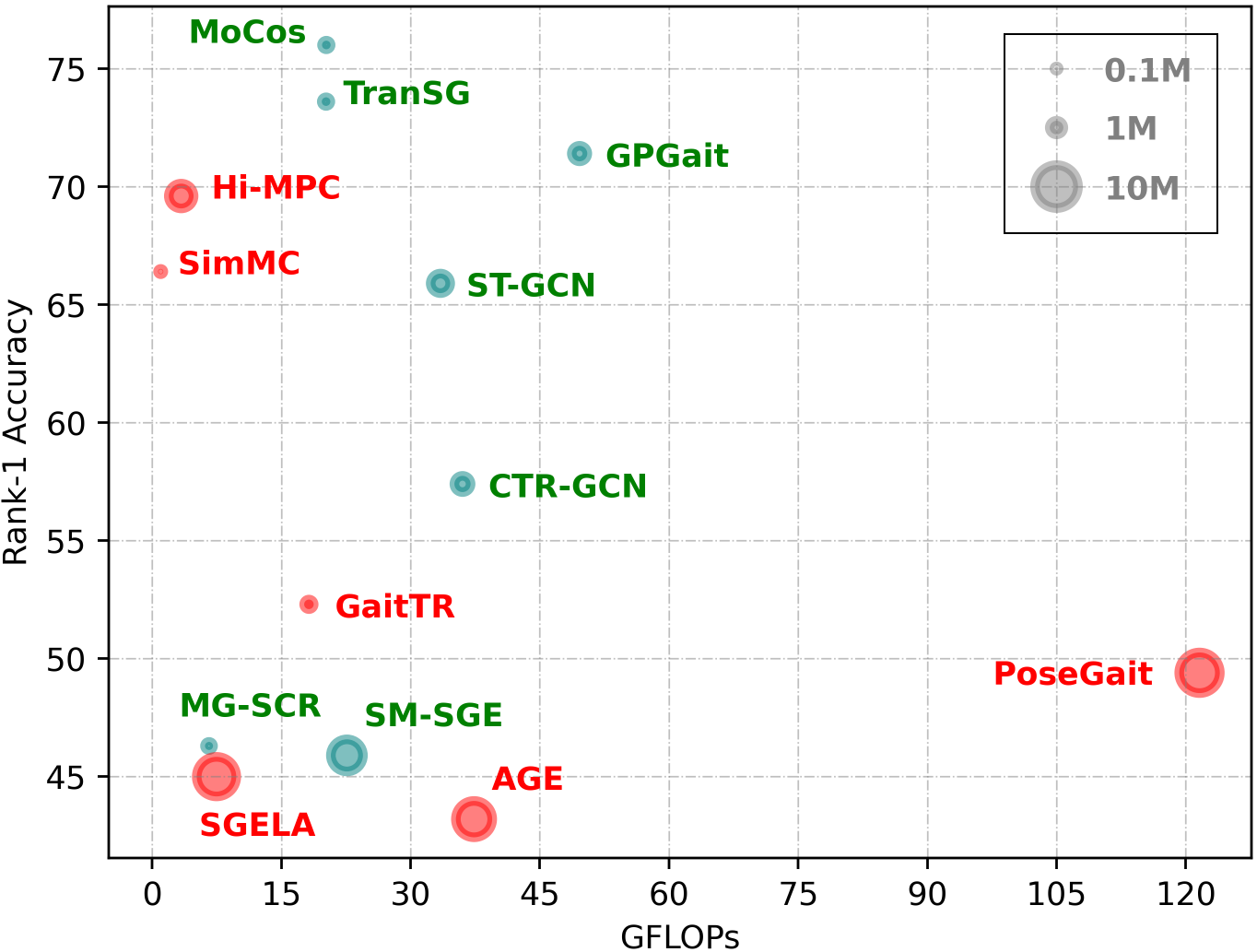}
     }
 	  \end{minipage}
 	  \label{model_comparison}
 	}
    \caption{(a) Overview of research origin and technical advancements of SRID within the person re-ID community (Zoom in and follow the timeline for the best view). (b) Parameter sizes (Millions (M)), computational complexity (Giga Floating Point Operations (GFLOPs)), and KS20 Rank-1 accuracy of state-of-the-art deep learning methods for SRID (\textcolor{red}{\textbf{Red}}: Sequence-based models; \textcolor{ForestGreen}{\textbf{Green}}: Graph-based models).}
    \label{xx}
\end{figure*}

In recent years, research on SRID has gained significant momentum, leading to diverse skeleton modeling and learning paradigms. Early endeavors  \cite{barbosa2012re,munaro20143d,andersson2015person,pala2019enhanced} mainly extract hand-crafted features such as skeleton descriptors in terms of anthropometric, geometric and gait attributes of body.
As these methods often require domain expertise such as anatomy and kinematics  \cite{yoo2002extracting} for skeleton modeling, they lack the ability to fully mine latent high-level features beyond human cognition.
 To resolve this challenge, recent mainstream methods \cite{liao2020model,huynh2020learning,rao2021self,rashmi2022human} leverage deep neural networks to automatically perform skeleton representation learning for SRID. One of exemplar methods (termed “\textit{sequence-based modeling}”) is to model sequential dynamics and motion semantics from raw or normalized skeletons ($e.g.,$ joint trajectory) based on long short-term memory (LSTM) and its variants \cite{wei2020person,rao2021self}. However, they rarely investigate the intrinsic body relationships such as inter-joint motion correlations, thereby possibly overlooking some valuable skeleton patterns. Another paradigm (termed “\textit{graph-based modeling}”) mitigates this challenge by constructing skeleton graphs to model discriminative structural and actional features based on the interrelations of body parts \cite{rao2023transg}. This often requires multi-granularity body modeling and efficient relational reasoning  mechanisms ($e.g.,$ collaborative learning) based on skeleton graphs.
Despite the great progress of SRID, this rapidly evolving technique still lacks a systematic review,  making it difficult for researchers to gain a holistic view of this field and embark on new research endeavors.

\begin{figure*}[t]
    \centering
    % \scalebox{0.231}{
\includegraphics[width=0.99\linewidth]{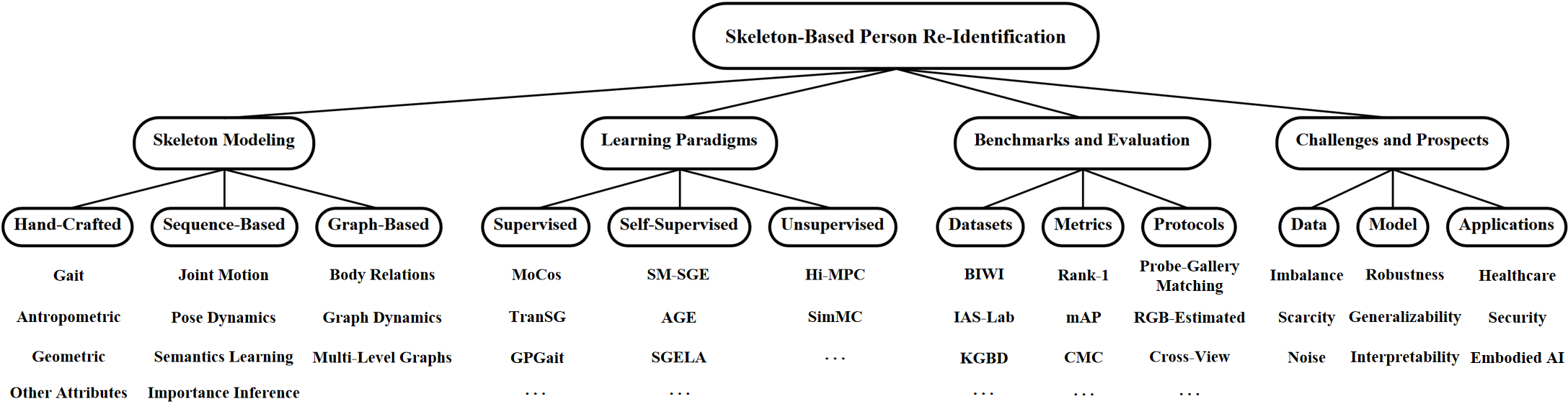}
    % }
    \caption{Structure of this survey with the taxonomy of SRID research. Representative branches and SRID methods are listed.} 
    \label{overview_SRID}
\end{figure*}

In light of this, we present the first survey on SRID, elucidating recent advancements of skeleton modeling, learning paradigms, evaluation benchmarks, current challenges, and interdisciplinary applications. Firstly, we define the SRID task and provide a milestone overview to illustrate the origin and key advancements of SRID as shown in Fig. \ref{SRID-Timeline}. Secondly,
we propose a systematic taxonomy of SRID methods to categorize them into hand-crafted, sequence-based, and graph-based modeling, and elaborate on their foundational mechanisms and representative approaches.
We also illustrate the basic definitions and common methods within three mainstream SRID paradigms (supervised, self-supervised, unsupervised).
Thirdly, we introduce existing public benchmarks, evaluation metrics, and protocols for SRID, while comprehensively evaluating state-of-the-art methods across different benchmarks to compare their performance and efficiency.
Meanwhile, we conduct qualitative analysis of different methods to compare their key properties with a discussion of advantages and disadvantages.
Finally, we discuss the current challenges in SRID and identify potential directions for future research.
An overview of promising SRID applications in interdisciplinary areas, spanning healthcare, embodied AI, and security, is further provided and illustrated with a case study.
The structure of this survey, including skeleton modeling (Sec. \ref{sec_skeleton_modeling}), learning paradigms (Sec. \ref{sec_learning_paradigms}), benchmarks and evaluation (Sec. \ref{sec_benchmark}), challenges and prospects (Sec. \ref{sec_challenges}) is shown in Fig. \ref{overview_SRID}. We hope our survey can bring new insights to researchers and expedite future research in SRID.

\section{Overview}
\label{sec_prelimiary}
\subsection{Task Description}
\label{sec_task_description}
As illustrated in Fig. \ref{task_overview}, the input of SRID task is a 3D skeleton sequence that belongs to a certain pedestrian, and the output is the predicted identity. 
Formally, we denote a 3D skeleton sequence as $\boldsymbol{S}\!=\!(\boldsymbol{s}_1,\cdots,\boldsymbol{s}_{f})\in \mathbb{R}^{f \times j \times 3}$, where $\boldsymbol{s}_{t}\in \mathbb{R}^{j \times 3}$ denotes the $t^{th}$ skeleton with 3D coordinates of $j$ body joints.
 Each skeleton sequence $\boldsymbol{S}$ corresponds to a person identity $\text{I}$, where $\text{I}\in \{1, \cdots, \text{C}\}$ and $\text{C}$ is the number of different classes ($i.e.$, identities). In the SRID task, we generally have \textit{training} set, \textit{probe} set, and \textit{gallery} set, respectively denoted as $\Phi_{T}=\left\{\boldsymbol{S}^{T}_{i}\right\}_{i=1}^{N_{1}}$, $\Phi_{P}=\left\{\boldsymbol{S}^{P}_{i}\right\}_{i=1}^{N_{2}}$, and $\Phi_{G}=\left\{\boldsymbol{S}^{G}_{i}\right\}_{i=1}^{N_{3}}$ that contain $N_{1}$, $N_{2}$, and $N_{3}$ skeleton sequences of different persons collected from different scenes or views. The task target is to learn a hand-crafted, sequence-based or graph-based model (detailed in Sec. \ref{sec_skeleton_modeling}) that maps 3D skeleton sequences into effective representations, so that we can query the correct identity of an encoded skeleton sequence representation in the probe set via matching it with the sequence representations in the database ($i.e.$, gallery set). SRID is essentially a retrieving and matching problem.

\subsection{Origin and Advancements}
As presented in Fig. \ref{SRID-Timeline}, the first SRID research \cite{munaro20143d} using hand-crafted skeleton descriptors commenced in 2014, coming after the first RGB video based method \cite{farenzena2010person} and before the first depth-based approach \cite{haque2016recurrent}. Then, Rao \textit{et al.} proposed the first deep learning based SRID paradigm in 2020, followed by the first self-supervised, unsupervised, and supervised paradigms \cite{rao2020self,rao2022simmc,rao2023transg}. Over the past five years, an increasing number of innovative models have been devised specifically for SRID and related emerging tasks, including LSTM models ($e.g.,$ AGE \cite{rao2020self}), CNN models ($e.g.,$ SkeletonGait \cite{fan2024skeletongait}), Transformer models ($e.g.$, TranSG \cite{rao2023transg}), GCN models ($e.g.,$ CTR-GCN \cite{chen2021channel}), MLP models ($e.g.$, SimMC \cite{rao2022simmc}), and hybrid/composite models ($e.g.$, SM-SGE \cite{rao2021sm}) (compared in Table \ref{performance_comparision}).

\subsection{Taxonomy of SRID Methods}
\label{sec_taxonomy}

As shown in Fig. \ref{overview_SRID}, we propose a systematic taxonomy for SRID approaches based on \textbf{skeleton modeling} (see Sec. \ref{sec_skeleton_modeling}) and \textbf{learning paradigms} (see Sec. \ref{sec_learning_paradigms}).
% \textbf{(1) Skeleton modeling} (see Sec. \ref{sec_skeleton_modeling}); \textbf{(2) Learning paradigms} (see Sec. \ref{sec_learning_paradigms}); 
% \textbf{(3) Benchmarks and evaluation} (see Sec. \ref{sec_benchmark});
% \textbf{(4) Challenges and directions} (see Sec. \ref{sec_challenges}). 
For skeleton modeling, we divide them into three categories, 
including (1) \textit{hand-crafted modeling} using manually-extracted features ($e.g.,$ skeleton descriptors), (2) \textit{sequence-based modeling} that focuses on sequential features ($e.g.,$ joint trajectory) of 3D skeletons, and (3) \textit{graph-based modeling} that represents 3D skeletons as graphs, and further subcategorize them by different learning focuses such as body relations or graph dynamics. 
In terms of learning paradigms, we group them into (1) \textit{supervised} SRID paradigms that require skeleton labels for feature learning, (2) \textit{self-supervised} SRID paradigms that combine pretext tasks for skeleton representation learning with labeled fine-tuning, and (3) \textit{unsupervised} SRID paradigms that learn skeleton features without using labels.

\section{Skeleton Modeling}
\label{sec_skeleton_modeling}
We elaborate on different skeleton modeling including their foundational mechanisms and representative approaches.

\subsection{Hand-Crafted Modeling}
\label{sec_hand}
\textbf{Gait Attributes.}
Extracting gait features is a common way to characterize unique walking patterns of an individual \cite{cunado2003automatic}, typically including (1) kinematic parameters ($e.g.,$ angles of hips, knees, and feet), and (2) spatio-temporal parameters ($e.g.,$ stride length, gait cycle time, velocity). They can be manually computed with:
\begin{equation}
    f_{angles} = \{(\alpha_{ij}, \beta_{ij}) \mid (i, j) \in \Psi\},
    \label{eq_angle_1}
\end{equation}
{\small
\begin{equation}
    \alpha_{ij} = \arctan\frac{y_i - y_j}{x_i - x_j}, \  \beta_{ij} = \arctan\frac{z_i - z_j}{\sqrt{(x_i - x_j)^2 + (y_i - y_j)^2}},
    \label{eq_angle_2}
\end{equation}}
where $x_{i}, y_{i}, z_{i}$ denote 3D coordinates of the $i^{th}$ joint, the set $\Psi$ defines \textit{adjacent} joints constrained by the human skeleton model, and two kinds of angles $\alpha_{ij}$ and $\beta_{ij}$ are calculated from these joint pairs. The velocity can be calculated by
\begin{equation}
    Velocity=\frac{\sum^{n}_{i=1}\frac{strideLength}{n}}{cycleTime},
    \label{gait_para_eq_1}
\end{equation}
where
\begin{equation}
    strideLength=2 * stepLength.
    \label{gait_para_eq_2}
\end{equation}
We calculate the step length by averaging the highest values of the
difference between the right and left feet, and adopt the mean stride length over all $n$ strides following \cite{andersson2015person}.

\textbf{Anthropometric Attributes.}
The Euclidean distance between two joints, such as bone lengths, limb dimensions, height, are usually computed as the anthropometric features ($f_{A}$) to differentiate individuals \cite{barbosa2012re}:
\begin{equation}
    f_{A} = \{\| {J}_i - {J}_j \|_2\, \mid (i, j) \in \Psi^{\ast}\},
    \label{eq_anthro}
\end{equation}
where ${J}_i\!\!=\!\!(x_{i}, y_{i}, z_{i})$, ${J}_j\!\!=\!\!(x_{j}, y_{j}, z_{j})$, $\Psi^{\ast}$ not only contains adjacent joints but also can be extended to cover more anthropometric properties, such as 13 ($D_{13}$) or 16 ($D_{16}$) skeleton descriptors in \cite{munaro2014one,pala2019enhanced}. 

\textbf{Geometric Attributes.} 
The incorporation of geometric skeletal measurements, including body-part ratios and inter-joint geodesic distances on the mesh surface, can  enhance feature representations in person re-ID \cite{barbosa2012re}.

The above hand-crafted features are often learned by different classifiers ($e.g.$, KNN) to perform person re-ID \cite{andersson2015person,nambiar2017context}. They are also combined with different metric algorithms \cite{pala2015multimodal} 
% % add revisit
or
other modalities such as 3D point clouds \cite{gharghabi2015people,bondi2018long,pala2019enhanced,munaro20143d} to further boost person re-ID accuracy.

\subsection{Sequence-Based Modeling}
\label{sec_sequence}
\textbf{Joint Motion.} 
The walking patterns are characterized by the motion of joints, which can be defined as the differences between body joint positions \cite{liao2020model}:
\begin{equation}
    f_{motion} = \boldsymbol{s}_{t} - \boldsymbol{s}_{t-1},
\end{equation}
where $s_{t}=\{J^{t}_{1}, J^{t}_{2},\cdots, J^{t}_{j}\}$, $J^{t}_{i}=(x^{t}_{i},y^{t}_{i},z^{t}_{i})$ denotes the 3D coordinates of $i^{th}$ joint in the $t^{th}$ skeleton, and $i\in\{1, 2, \cdots,j\}$. 
The occluded or masked joints during walking can also help models learn useful motion patterns \cite{rashmi2022human}.

\textbf{Pose Dynamics.} 
The consecutive skeletons typically conveys dynamics of unique body poses, which can be encoded sequentially by temporal learning models ($e.g.,$ LSTM):
\begin{equation}
\boldsymbol{h}_t = 
\begin{cases} 
\phi(\boldsymbol{s}_1) & \text{if } t = 1 \\ 
\phi(\boldsymbol{h}_{t-1}, \boldsymbol{s}_t) & \text{if } 1< t \leq f 
\end{cases},
\end{equation}
where $\phi(\cdot)$ denotes the model function to encode the pose dynamics of skeletons, $\boldsymbol{h}_{t-1}$ represents the latent representation of previous $(t-1)$ poses, which provides the temporal context information to encode the long-term pose dynamics $\boldsymbol{h}_{t}$ for person re-ID \cite{rao2021self}. The joint distances, relative joint positions, and bones are combined in \cite{wei2020person,zhang2023spatial} to enhance pose dynamics learning.

\textbf{Semantics Learning.}
By encoding the skeleton sequence into latent high-dimensional representation, latent motion semantics such as motion continuity can be captured to enhance feature learning.
Representative semantics learning tasks include skeleton sequence reconstruction and prediction, which can be simply formulated as follows:
\begin{equation}
\phi(\boldsymbol{s}_1,\boldsymbol{s}_2,\cdots,\boldsymbol{s}_f)=
\begin{cases} 
\boldsymbol{s}_1,\boldsymbol{s}_2,\cdots,\boldsymbol{s}_f & \text{Plain Recon.}  \\ 
\boldsymbol{s}_f,\boldsymbol{s}_{f-1},\cdots,\boldsymbol{s}_1 & \text{Reverse Recon.}  \\
\boldsymbol{s}_{f+1},\boldsymbol{s}_{f+2},\cdots,\boldsymbol{s}_{2f} & \text{Prediction}
\end{cases}
\label{eq_recon_pred}
\end{equation}
The reverse reconstruction (Recon.) and prediction require the model function $\phi(\cdot)$ to capture high-level semantics ($e.g.,$ order and correlations) to achieve the target output, which facilitates learning more meaningful gait representations for person re-ID \cite{rao2021self}. 
A few studies also explore the semantics learning of motion consistency \cite{rao2022simmc} and  cross-scale body relations \cite{rao2021sm}.

\textbf{Importance Inference.}
Different skeletons and their feature representations typically possess different importance in characterizing poses and discriminative patterns of a person, which can be exploited to mine key skeletons or hard samples \cite{hermans2017defense} for SRID learning.
To aggregate key skeleton features for SRID,  AGE \cite{rao2020self} explores a locality-aware attention mechanism to integrate features of important skeletons in the sequence, while SM-SGE \cite{rao2021sm} infers key correlations between different body-joint nodes. Rao \textit{et al.} \shortcite{rao2024hierarchical} further propose a hard skeleton mining mechanism to adaptively infer importance of multi-level skeleton representations for SRID.

\subsection{Graph-Based Modeling}
\label{sec_graph}
\textbf{Body Relations.}
Skeletons can be naturally modeled as graphs based on the physical connections of human body
joints. For each skeleton $\boldsymbol{x}_t$, we represent it as the graph \( \mathcal{G}^t (\mathcal{V}^t, \mathcal{E}^t) \), where  
$\mathcal{V}^t = \{ v_1^t, v_{2}^t, \cdots, v_{j}^t \}, v_i^t \in \mathbb{R}^3, i \in \{ 1, \cdots, j \}$ 
and edges  $\mathcal{E}^t = \{ e_{i,j}^t \mid v_i^t, v_j^t \in \mathcal{V}^t \}, e_{i,j}^t \in \mathbb{R}$.  
$ \mathcal{E}^t $ denotes the set of connections between different joints.
Based on the skeleton graphs, we can model the inherent correlations, such as limb collaboration \cite{rao2021multi,rao2022skeleton} and channel-specific relations \cite{chen2021channel} for SRID. The graph motifs can also be devised to enhance the structure and gait related relation learning \cite{rao2025motif}.

\textbf{Graph Dynamics.}
The dynamics of skeleton graphs can be leveraged to learn and capture the temporal evolution of joints' connection patterns and limbs' coordination for SRID. The process can be simplified as
\begin{equation}
\boldsymbol{g}_t = 
\begin{cases} 
\phi\left(f_{G}\left(\boldsymbol{s}_1\right)\right) & \text{if } t = 1 \\ 
\phi\left(\boldsymbol{g}_{t-1}, f_{G}\left(\boldsymbol{s}_t\right)\right) & \text{if } 1< t \leq f 
\end{cases},
\end{equation}
where $f_{G}(\cdot)$ denotes the graph encoding model ($e.g.,$ graph convolutional networks \cite{fu2023gpgait}), $\phi(\cdot)$ is the temporal learning model ($e.g.,$ LSTM) to encode the long-term dynamics of graph representations. To facilitate spatio-temporal graph learning, self-supervised pretext tasks, such as graph reconstruction and prediction, are incorporated to learn more effective skeleton features \cite{rao2023transg}. 

\textbf{Multi-Level Graphs.} 
 Multi-level skeleton graph representations are devised to characterize coarse-to-fine body structure and motion \cite{li2020dynamic}, and various graph learning tasks such as sparse graph prediction and cross-scale graph inference \cite{rao2021sm} are proposed to help learn different-level graph semantics. Based on graph transformers, masked reconstruction \cite{he2022masked} and motifs \cite{rao2025motif} are further explored for high-level structural ($e.g.,$ locality) and gait semantics learning.

\begin{table}[t]
\centering
\scalebox{0.6}{
\renewcommand\arraystretch{1.5}{
% 4.4
\setlength{\tabcolsep}{0.9mm}{
\begin{tabular}{cllcccc}
\hline
\textbf{Type} & \textbf{Dataset} & \textbf{Reference} & \textbf{Source} & \textbf{\# ID} & \textbf{\# Skeletons} & \textbf{\# View} \\ \hline
\multirow{8}{*}{\rotatebox{90}{\textbf{Sensor-Based}}} & BIWI RGBD-ID & \cite{munaro2014one} & Kinect V1 & 50 & 205.8K & S \\
 & IAS-Lab RGBD-ID & \cite{munaro2014feature} & Kinect V1 & 11 & 89.0K & S \\
 & KGBD & \cite{andersson2015person} & Kinect V1 & 164 & 188.7K & S \\
 & KinectREID & \cite{pala2015multimodal} & Kinect V1 & 71 & 4.8K & 7 \\
 & UPCV1 & \cite{kastaniotis2015framework} & Kinect V1 & 30 & 13.1K & S \\
 & UPCV2 & \cite{kastaniotis2016gait} & Kinect V2 & 30 & 26.3K & S \\
 & Florence 3D Re-ID & \cite{bondi2018long} & Kinect V2 & 16 & 18.0K & S \\
 & KS20 & \cite{nambiar2017context} & Kinect V2 & 20 & 36.0K & 5 \\ \hline
\multirow{4}{*}{\rotatebox{90}{\textbf{Estimated}}} & CAISA-B-3D & \cite{liao2020model} & Videos & 124 & 706.5K & 11 \\
 & 3DGait$^{\star}$ & \cite{wang2023amai} & Videos & 43 & 22.9K & S \\
 & OUMVLP-Pose-2D & \cite{chen2022keypoint} & Videos & 10307 & 6667.0K & 14 \\
 & PoseTrackReID-2D & \cite{chen2022keypoint} & Videos & 5350 & 53.6K & — \\ \hline
\end{tabular}
}
}
}
\caption{Overview of SRID benchmark datasets. The number of skeletons in training is reported. “S” denotes single or egocentric view. \textit{RGB-estimated} 3D and 2D skeleton datasets are listed. $^{\star}$ denotes an interdisciplinary benchmark for gait and disease prediction.}
\label{dataset_statistics}
\end{table}

\begin{table*}[t]
\scalebox{0.6}{
\renewcommand\arraystretch{1.3}{
\setlength{\tabcolsep}{0.9mm}{
\begin{tabular}{llccccccccccccc}
\hline
\textbf{} & \textbf{Method} & \textbf{\begin{tabular}[c]{@{}c@{}}Conference/\\ Journal\end{tabular}} & \textbf{BIWI-S} & \textbf{BIWI-W} & \textbf{IAS-A} & \textbf{IAS-B} & \textbf{KS20} & \textbf{\begin{tabular}[c]{@{}c@{}}Algorithms/\\ Architectures\end{tabular}} & \textbf{\begin{tabular}[c]{@{}c@{}}Network\\ Parameters\end{tabular}} & \textbf{\begin{tabular}[c]{@{}c@{}}Require\\ Labels?\end{tabular}} & \textbf{\begin{tabular}[c]{@{}c@{}}Pretext\\ Task?\end{tabular}} & \textbf{\begin{tabular}[c]{@{}c@{}}Multi-Scale\\ Modeling?\end{tabular}} & \textbf{\begin{tabular}[c]{@{}c@{}}Theory for \\ Analysis?\end{tabular}} & \textbf{Summary} \\ \hline
\multirow{9}{*}{\textbf{\rotatebox{90}{Hand-Crafted}}} & \cite{gharghabi2015people} & ICIEV & 10.7 & 10.7 & — & — & — & KNN & — & \textcolor{ForestGreen}{\checkmark} & \textcolor{red}{\ding{55}} & \textcolor{red}{\ding{55}} & \textcolor{ForestGreen}{\checkmark} & \multirow{3}{*}{\begin{tabular}[c]{@{}c@{}}\textbf{Advantages:}\\ High theoretical explainability \\ of features and models;\\ Generally low complexity.\end{tabular}} \\
 & \cite{munaro20143d} & \begin{tabular}[c]{@{}c@{}}Person \\ Re-Identification\end{tabular} & 32.1 & 39.3 & — & — & — & NN & — & \textcolor{ForestGreen}{\checkmark} & \textcolor{red}{\ding{55}} & \textcolor{red}{\ding{55}} & \textcolor{ForestGreen}{\checkmark} &  \\
 & $D_{13} \ $\cite{munaro2014one} & ICRA & 28.3 & 14.2 & 40.0 & 43.7 & 39.4 & NN & — & \textcolor{ForestGreen}{\checkmark} & \textcolor{red}{\ding{55}} & \textcolor{red}{\ding{55}} & \textcolor{ForestGreen}{\checkmark} &  \\
 & $D_{16}$ \cite{pala2019enhanced} & \begin{tabular}[c]{@{}c@{}}Computers\&\\ Graphics\end{tabular} & 32.6 & 17.0 & 42.7 & 44.5 & 51.7 & Adaboost & — & \textcolor{ForestGreen}{\checkmark} & \textcolor{red}{\ding{55}} & \textcolor{red}{\ding{55}} & \textcolor{ForestGreen}{\checkmark} & \multirow{3}{*}{\begin{tabular}[c]{@{}c@{}}\textbf{Disadvantages:}\\ Require domain knowledge;\\ Labor-expensive;\\ Low performance.\end{tabular}} \\
 & \cite{elaoud2017analysis} & ACIVS & 28.6 & 10.7 & 45.5 & 63.6 & — & SVD & — & \textcolor{ForestGreen}{\checkmark} & \textcolor{red}{\ding{55}} & \textcolor{red}{\ding{55}} & \textcolor{ForestGreen}{\checkmark} &  \\
 & PM \cite{elaoud2021person} & \begin{tabular}[c]{@{}c@{}}Multimedia Tools \\ and Applications\end{tabular} & 39.3 & 39.3 & 36.4 & 81.8 & — & RF & — & \textcolor{ForestGreen}{\checkmark} & \textcolor{red}{\ding{55}} & \textcolor{red}{\ding{55}} & \textcolor{ForestGreen}{\checkmark} &  \\ \hline
\multirow{6}{*}{\textbf{\rotatebox{90}{Sequence-Based}}} & PoseGait \cite{liao2020model} & Pattern Recognition & 14.0 & 8.8 & 28.4 & 28.9 & 49.4 & CNN & 8.93M & \textcolor{ForestGreen}{\checkmark} & \textcolor{red}{\ding{55}} & \textcolor{red}{\ding{55}} & \textcolor{red}{\ding{55}} & \multirow{3}{*}{\begin{tabular}[c]{@{}c@{}}\textbf{Advantages:}\\ Learn with raw sequences;\\ Capture temporal dynamics.\end{tabular}} \\
 & AGE \cite{rao2020self} & IJCAI & 25.1 & 11.7 & 31.1 & 31.1 & 43.2 & LSTM & 7.15M & \textcolor{ForestGreen}{\checkmark} & \textcolor{ForestGreen}{\checkmark} & \textcolor{red}{\ding{55}} & \textcolor{red}{\ding{55}} &  \\
 & SGELA  \cite{rao2021self} & TPAMI & 25.8 & 11.7 & 16.7 & 22.2 & 45.0 & LSTM & 8.47M & \textcolor{ForestGreen}{\checkmark} & \textcolor{ForestGreen}{\checkmark} & \textcolor{red}{\ding{55}} & \textcolor{red}{\ding{55}} &  \\
 & GaitTR \cite{zhang2023spatial} & Expert Systems & 43.2 & 16.3 & 43.7 & 47.8 & 52.3 & Transformer & 0.49M & \textcolor{ForestGreen}{\checkmark} & \textcolor{red}{\ding{55}} & \textcolor{red}{\ding{55}} & \textcolor{red}{\ding{55}} & \multirow{3}{*}{\begin{tabular}[c]{@{}c@{}}\textbf{Disadvantages:}\\ Lack body structure modeling;\\ Ignore joint or limb relations.\end{tabular}} \\
 & SimMC \cite{rao2022simmc} & IJCAI & 41.7 & 24.5 & 44.8 & 46.3 & 66.4 & MLP & 0.15M & \textcolor{red}{\ding{55}} & \textcolor{red}{\ding{55}} & \textcolor{red}{\ding{55}} & \textcolor{ForestGreen}{\checkmark} &  \\
 & Hi-MPC \cite{rao2024hierarchical} & IJCV & 47.5 & 27.3 & 45.6 & 48.2 & 69.6 & MLP & 3.32M & \textcolor{red}{\ding{55}} & \textcolor{red}{\ding{55}} & \textcolor{ForestGreen}{\checkmark} & \textcolor{ForestGreen}{\checkmark} &  \\ \hline
\multirow{8}{*}{\textbf{\rotatebox{90}{Graph-Based}}} & SkeletonGait \cite{fan2024skeletongait} & AAAI & 15.1 & 10.8 & 31.4 & 31.5 & 22.2 & CNN & 11.11M & \textcolor{ForestGreen}{\checkmark} & \textcolor{red}{\ding{55}} & \textcolor{red}{\ding{55}} & \textcolor{red}{\ding{55}} & \multirow{5}{*}{\begin{tabular}[c]{@{}c@{}}\textbf{Advantages:}\\ Fully model body structure\\ and motion connections;\\ Improved interpretability\\ with relations.\end{tabular}} \\
 & MG-SCR \cite{rao2021multi} & IJCAI & 20.1 & 10.8 & 36.4 & 32.4 & 46.3 & GAT, LSTM & 0.35M & \textcolor{ForestGreen}{\checkmark} & \textcolor{ForestGreen}{\checkmark} & \textcolor{ForestGreen}{\checkmark} & \textcolor{red}{\ding{55}} &  \\
 & SM-SGE \cite{rao2021sm} & ACM MM & 31.3 & 13.2 & 34.0 & 38.9 & 45.9 & MGRN, LSTM & 5.58M & \textcolor{ForestGreen}{\checkmark} & \textcolor{ForestGreen}{\checkmark} & \textcolor{ForestGreen}{\checkmark} & \textcolor{red}{\ding{55}} &  \\
 & CTR-GCN \cite{chen2021channel} & CVPR & 59.1 & 20.5 & 47.7 & 48.3 & 57.4 & GCN & 1.42M & \textcolor{ForestGreen}{\checkmark} & \textcolor{red}{\ding{55}} & \textcolor{red}{\ding{55}} & \textcolor{red}{\ding{55}} &  \\
 & ST-GCN \cite{yan2018spatial} & AAAI & 56.8 & 21.3 & 47.9 & 50.1 & 65.9 & GCN & 2.06M & \textcolor{ForestGreen}{\checkmark} & \textcolor{red}{\ding{55}} & \textcolor{red}{\ding{55}} & \textcolor{red}{\ding{55}} &  \\
 & GPGait \cite{fu2023gpgait} & ICCV & 54.1 & 29.0 & 50.9 & 60.1 & 71.4 & GCN & 1.30M & \textcolor{ForestGreen}{\checkmark} & \textcolor{red}{\ding{55}} & \textcolor{red}{\ding{55}} & \textcolor{red}{\ding{55}} & \multirow{3}{*}{\begin{tabular}[c]{@{}c@{}}\textbf{Disadvantages:}\\ Graph topology pre-defining;\\ Complex relation modeling.\end{tabular}} \\
 & TranSG \cite{rao2023transg} & CVPR & 68.7 & 32.7 & 49.2 & 59.1 & 73.6 & Transformer & 0.40M & \textcolor{ForestGreen}{\checkmark} & \textcolor{ForestGreen}{\checkmark} & \textcolor{red}{\ding{55}} & \textcolor{ForestGreen}{\checkmark} &  \\
 & MoCos \cite{rao2025motif} & AAAI & 72.0 & 36.0 & 51.9 & 61.5 & 76.0 & Transformer & 0.40M & \textcolor{ForestGreen}{\checkmark} & \textcolor{ForestGreen}{\checkmark} & \textcolor{red}{\ding{55}} & \textcolor{ForestGreen}{\checkmark} &  \\ \hline
\end{tabular}
}
}
}
\caption{Performance (R$_{1}$) and characteristics comparison of existing hand-crafted, sequence-based, and graph-based methods on different benchmark datasets (BIWI (S/W), IAS-Lab (A/B), KS20). Representative gait recognition methods using skeleton data are also compared following the same person re-ID evaluation protocol. We report the parameter sizes (\textbf{M}illion) and summarize their properties.}
\label{performance_comparision}
\end{table*}

\section{Learning Paradigms}
\label{sec_learning_paradigms}

   \textbf{Supervised} SRID paradigms leverage skeletal annotations or labels ($e.g.,$ identity class) to guide the model to learn discriminative features, typically using cross-entropy (CE) loss as follows:
\begin{equation}
    \mathcal{L}_{\text{CE}} = \frac{1}{N_{1}}\sum_{i=1}^{N_{1}}\sum_{j=1}^{C} -\text{I}_{i,j} \log(\hat{\text{I}}_{i,j}),
    \label{eq_cross_entropy}
\end{equation}
    where $\text{I}_{i,j}$ denotes the ground-truth identity label  ($\text{I}_{i,j}=1$ iff the $i^{th}$ sample belongs to the $j^{th}$ identity otherwise 0), $\hat{\text{I}}_{i,j}$ is the probability that the $i^{th}$ sample is predicted as the $j^{th}$ identity, and $N_{1}$ is the number of training samples. This loss has been widely applied to classic supervised models such as SVM \cite{munaro2014one}, KNN \cite{munaro20143d,gharghabi2015people,nambiar2017context}, MLP \cite{andersson2015person}, Adaboost \cite{pala2019enhanced}, random forest \cite{elaoud2021person} to learn skeleton descriptors to classify different individuals. Recent studies explore supervised skeleton prototype learning paradigms by employing the contrastive loss
\begin{equation}
    \mathcal{L}_{\text{Proto}}=\frac{1}{N_{1}} \sum_{k=1}^{C} \sum_{j=1}^{n_k}-\log \frac{\exp \left(\boldsymbol{S}_{k, j} \cdot \boldsymbol{p}_{k} / \tau\right)}{\sum_{i=1}^{C} \exp \left(\boldsymbol{S}_{k, j} \cdot \boldsymbol{p}_{i} / \tau\right)},
    \label{eq_contrast}
\end{equation}
where $\boldsymbol{S}_{k, j}$ denotes the $j^{th}$ skeleton sequence belonging to the $k^{th}$ identity, $\boldsymbol{p}_{k}$ represents the prototype generated by the feature centroid of $k$-class skeletons, and $\tau$ is the temperature for contrastive learning. 
Based on the losses in Eq. (\ref{eq_cross_entropy}) and (\ref{eq_contrast}), diverse network architectures such as skeleton-based LSTM \cite{wei2020person,rashmi2022human}, CNN \cite{liao2020model,huynh2020learning}, and Transformer \cite{rao2023transg} are trained for SRID.

\textbf{Self-Supervised} SRID paradigms usually combine unlabeled learning of pretext tasks ($e.g.,$ skeleton semantics learning) and labeled fine-tuning. A general form of pretext objective can be formulated as
\begin{equation}
\mathcal{L}_{\text{Pretext}} = \frac{1}{N_1} \sum_{i=1}^{N_1} \text{Dis}\left( \phi\left(T_{1}\left({\boldsymbol{S}}_i\right)\right), T_{2}({\boldsymbol{S}}_i)\right),
\end{equation}
where $T_{1}(\cdot)$, $T_{2}(\cdot)$ denote the sequence transformation functions ($e.g.,$ reverse and shift in Eq. (\ref{eq_recon_pred})) to set the input and target of a pretext task, $\phi(\cdot)$ represent the model encoding function, and $\text{Dis}(\cdot)$ is the distance metric ($e.g.,$ Euclidean distance). By employing sequence-based pretext tasks such as sparse sequential prediction \cite{rao2021multi}, or graph-based tasks such as structure-trajectory prompted reconstruction \cite{rao2023transg}, self-supervised learning encourages the model to capture high-level motion concepts and class-related spatio-temporal semantics for SRID.

\textbf{Unsupervised} SRID paradigms perform skeleton representation learning without using any labels. Existing methods mainly adopt unsupervised skeleton prototype learning frameworks, which use the feature centroids of clustering as prototypes (see Eq. (\ref{eq_contrast})) \cite{rao2022simmc,rao2024hierarchical}. 
 Their performance could be affected by the robustness of clustering algorithms and contrastive learning mechanisms, while unlabeled pretext tasks are often combined to enhance general skeleton semantics learning.

\section{Benchmarks and Evaluation}
\label{sec_benchmark}

\begin{table}[t]
\centering
\scalebox{0.6}{
\renewcommand\arraystretch{1.4}{
\setlength{\tabcolsep}{2.3mm}{
\begin{tabular}{l|rrrrrrrrrr}
\hline
\textbf{Probe-Gallery} & \multicolumn{2}{c}{\textbf{N-N}} & \multicolumn{2}{c}{\textbf{B-B}} & \multicolumn{2}{c}{\textbf{C-C}} & \multicolumn{2}{c}{\textbf{C-N}} & \multicolumn{2}{c}{\textbf{B-N}} \\ \hline
\textbf{Methods} & \textbf{mAP} & \textbf{R$_{1}$} & \textbf{mAP} & \textbf{R$_{1}$} & \textbf{mAP} & \textbf{R$_{1}$} & \textbf{mAP} & \textbf{R$_{1}$} & \textbf{mAP} & \textbf{R$_{1}$} \\ \hline
AGE$^{\dagger}$ & 3.5 & 20.8 & 9.8 & 37.1 & 9.6 & 35.5 & 3.0 & 14.6 & 3.9 & 32.4 \\
SM-SGE$^{\clubsuit}$  & 6.6 & 50.2 & 9.3 & 26.6 & 9.7 & 27.2 & 3.0 & 10.6 & 3.5 & 16.6 \\
SPC-MGR$^{\clubsuit}$ & 9.1 & 71.2 & 11.4 & 44.3 & 11.8 & 48.3 & 4.3 & 22.4 & 4.6 & 28.9 \\
SGELA$^{\dagger}$ & 9.8 & 71.8 & 16.5 & 48.1 & 7.1 & 51.2 & 4.7 & 15.9 & 6.7 & 36.4 \\
SimMC$^{\dagger}$ & 10.8 & 84.8 & 16.5 & 69.1 & 15.7 & 68.0 & 5.4 & 25.6 & 7.1 & 42.0 \\
TranSG$^{\clubsuit}$ & 13.1 & 78.5 & 17.9 & 67.1 & 15.7 & 65.6 & 6.7 & 23.0 & 8.6 & 44.1 \\
Hi-MPC$^{\dagger}$ & 11.2 & 85.5 & 17.0 & 71.2 & 14.1 & 70.2 & 4.9 & 27.2 & 7.5 & 50.1 \\
MoCos$^{\clubsuit}$ & 16.1 & 87.9 & 18.9 & 73.6 & 18.1 & 72.1 & 7.3 & 26.5 & 9.8 & 50.6 \\ \hline
\end{tabular}
}
}
}
\caption{Performance of state-of-the-art SRID methods on different conditions  (\textbf{N}ormal, \textbf{C}lothes, \textbf{B}ags) of RGB-estimated CASIA-B dataset. ``\textbf{C-N}'' denotes ``Clothes'' probe set and ``Normal'' gallery set. $^{\dagger}$ and $^{\clubsuit}$ denote sequence-based and graph-based methods.}
\label{CASIA_comparision}
\end{table}

\textbf{Benchmark Datasets.} Table \ref{dataset_statistics} summarizes the data source, identity number, skeleton amount, and viewpoint number of commonly-used datasets for SRID evaluation. They can be mainly categorized to two types: (1) \textbf{Sensor-based datasets}, where 3D skeleton data are captured from depth sensors such as Kinect, and (2) \textbf{RGB-estimated datasets}, in which skeleton data are estimated from RGB videos using 2D or 3D pose estimation models \cite{chen20173d,cao2019openpose}. Existing SRID datasets typically contain skeleton data collected from varying scenarios such as multiple views (KS20, KinectREID), appearance and clothing changes (BIWI RGBD-ID, IAS-Lab RGBD-ID), and different illumination conditions (KGBD), which enables a comprehensive evaluation of both short-term and long-term SRID performance. 
In addition to standard SRID datasets, the 3DGait dataset provides an \textbf{interdisciplinary benchmark} for evaluating the generalizability of SRID models on healthcare-related tasks, such as neurodegenerative disease prediction (see Sec. \ref{sec_applications}) \cite{rao2025llm}.

\textbf{Evaluation Metrics and Protocols.} In SRID, the performance is typically evaluated based on several mertics, including Cumulative Matching Characteristics (CMC), Rank-1 accuracy (R$_{1}$), Rank-5 accuracy, Rank-10 accuracy, and Mean Average Precision (mAP) \cite{zheng2015scalable}. Multiple evaluation protocols are employed across varying datasets, including probe-gallery matching evaluation (main protocol evaluated in Table \ref{performance_comparision}), RGB-estimated evaluation (representative protocol evaluated in Table \ref{CASIA_comparision}), random view evaluation, cross-view
evaluation \cite{rao2024hierarchical}, zero-shot cross-dataset evaluation \cite{rao2021self}, etc.

\textbf{Comparison of Performance and Efficiency.}
As shown in Table \ref{performance_comparision},
$D_{16}$ and PM are two most competitive hand-crafted methods, performing well on both BIWI and IAS benchmarks.
However, recent deep learning based models significantly surpass them:
The latest sequence-based model (Hi-MPC) and graph-based model (MoCos) achieve superior performance to them across different datasets.
Notably, the top three performers—GPGait, TranSG, and MoCos—all utilize skeleton graph representations, highlighting the efficacy of graph modeling and body relation learning for SRID task.
In terms of efficiency ($cf.$ Fig. \ref{model_comparison}), the sequence-based SimMC demonstrates the most compact design requiring only 0.15M parameters, followed by the graph-based MG-SCR (0.32M).
Generally, methods utilizing Transformer architectures (GaitTR, TranSG, MoCos) exhibit significantly smaller parameter sizes than using other architectures such as LSTM. It is noteworthy that CNN-based approaches such as PoseGait incur substantially higher computational cost.

\textbf{Comparison of Key Characteristics.}
As summarized in Table \ref{performance_comparision}, hand-crafted methods design features based on domain knowledge, and employ well-established machine learning models, offering relatively high explainability and low model complexity. 
By contrast, sequence-based methods learn directly from raw data to capture sequential dynamics, but often overlook valuable features within body structure and joint relations.
Graph-based approaches are designed to model these structural and motion connections via pre-defined topologies. By quantifying joint importance and collaboration, graph models provide a more interpretable framework.
While both sequence-based and graph-based paradigms can integrate pretext tasks ($e.g.,$ reconstruction) and multi-scale modeling ($e.g.,$ multi-level body representations) to boost learning, only 4 out of 14 (28.6\%) of these methods provide theoretical analyses of their effectiveness.

\section{Challenges and Prospects}
\label{sec_challenges}
In this section, we elucidate key challenges in SRID data and models with a discussion of potential directions. We also present promising interdisciplinary applications of SRID.

\label{sec_data_challenges}
\textbf{Data Scarcity and Imbalance.} Existing 3D skeletons are mainly collected from prevailing depth sensors such as Kinect \cite{shotton2011real-time}, while diverse skeleton collection settings ($e.g.,$ different devices in uncontrollable environments) have not be thoroughly explored. In contrast to existing RGB-based person re-ID data ($e.g.,$ MSMT17 \cite{wei2018person}), the available skeleton data for person re-ID are relatively scarce and imbalanced. As shown in Table \ref{dataset_statistics}, existing Kinect-based SRID datasets contain 4.8 to 205.8 thousand skeletons with less than 200 identities, while the numbers of skeletons belonging to each identity often differ greatly ($i.e.,$ imbalanced class distribution). Training on these datasets might incur limited generalization ability of data-driven SRID models using deep neural networks, with a risk of model over-fitting.

\textbf{Data Noise.} Kinect-based and RGB-estimated skeleton data possibly contain noise due to: (1) Internal factors: The key points are generated by device built-in algorithms (via depth images) or pose estimation models (via RGB images), thus the inherent limitation of algorithms ($e.g.,$ precision, robustness) might result in inaccurate skeletons; (2) External factors:
The quality of skeleton data may also be affected by device's tracking distance, illumination changes ($e.g.,$ influence structured light in Kinect V1), and source data quality ($e.g.,$ image resolution).
Such inherent noise puts high demand on the model robustness against random perturbations.

To address these two challenges, higher-quality 3D skeleton data should be collected or generated. 
It is feasible to devise skeleton denoising models and augmentation strategies for skeleton generation. 
GAN-based pose generators \cite{yan2017skeleton} and diffusion models \cite{ho2020denoising} could be transferred to generate and denoise 3D skeleton data.
The future efforts include (a) collecting and opening new larger-scale SRID datasets, (b) transferring existing skeleton datasets from other areas to person re-ID datasets and formulating appropriate evaluation protocols, (c) estimating 2D/3D skeletons from large-scale public person re-ID datasets to construct new estimated SRID datasets, so as to advance SRID research and related person re-ID community.

\begin{figure*}
    \centering
\includegraphics[width=0.95\linewidth]{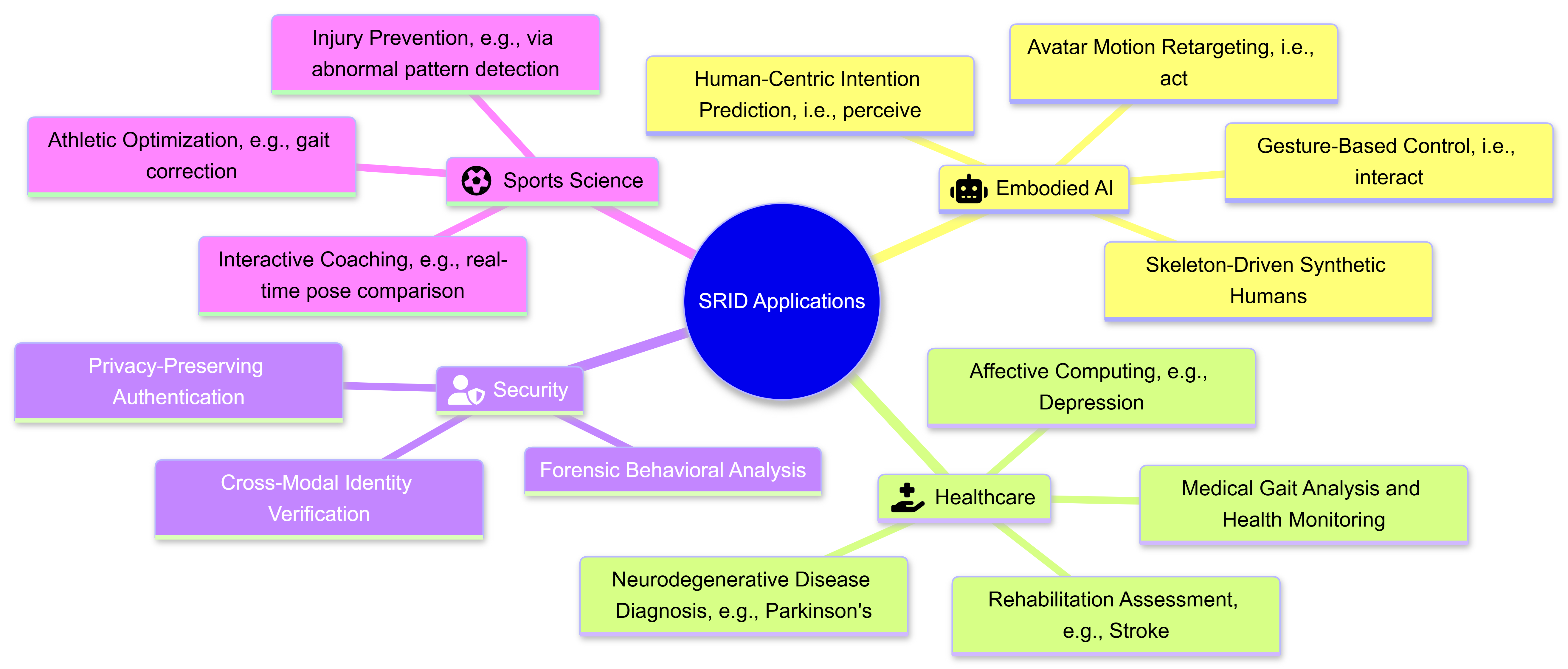}
    \caption{Interdisciplinary application landscape of SRID across three primary domains: healthcare ({green box}), embodied AI ({yellow box}), security ({purple box}), and sports science ({pink box}). Please zoom in for better view.}
    \label{application_overview}
\end{figure*}

\textbf{Model Robustness.} 
A few SRID models report unstable performance variations that are sensitive to different model parameter initialization, hyper-parameter settings ($e.g.,$ clustering parameters) and varying data distributions \cite{rao2022simmc}. As these studies opt for the best-trained model with finely-tuned parameters, this practice often fails to reflect the architecture's true average performance and shows limited adaptability in real-world applications.

A key future direction is to systematically investigate how key factors ($e.g.$, model initialization, data quality) affect robustness, which is essential for developing more reliable models.
Theoretical analyses of performance variations in terms of model-approximated functions and convergence conditions can also be provided for more robust model design. 
Moreover, multi-faceted evaluation metrics, such as performance average and standard deviation, should be reported to better measure the overall robustness of models. 

\textbf{Model Generalizability.} 
Most SRID models are trained on a single dataset with limited data sizes, views, scenes or conditions.
As a result, they can hardly generalize to different data domains ($e.g.,$ vulnerable to domain shifts) or real-world scenarios ($e.g.,$ RGB-estimated skeleton data). 

A potential solution is to exploit larger-scale SRID datasets to train the model across diverse scenarios, so as to learn more domain-general skeleton semantics. 
It is also feasible to explore domain adaptation or generalization techniques to co-train and transfer models. More benchmarks should be investigated for generality evaluation of SRID models.

\textbf{Interpretability.} 
Current SRID models typically lack intuitive explanations for the effectiveness of model architectures, skeleton features, and prediction results. This opacity not only poses risks of erroneous outcomes but also hinders their reliable applications at scale.

To this end, different human-friendly explanation including pose/feature visualization and corresponding language-based description can be considered. There also exist various architecture-specific explanation mechanisms, $e.g.,$ class activation maps (CAM) \cite{zhou2016learning} for CNN, knowledge graphs \cite{ji2021survey} for graph neural networks, which could be applied to explainable skeleton learning. 
Moreover, large language models (LLMs) can serve as an agent to co-train the SRID model  and interpret the learning process. A possible direction is to transform skeletons into time series text or pose images, and leverage LLMs or other agents to analyze the pose/feature importance.

\subsection{Emerging and Open Directions}
\label{sec_discussions}
% \begin{itemize}
    \textbf{Multi-Modal Learning.} Combining 3D skeletons and emerging modalities such as images \cite{bao2025vid}, radio frequency waves ($e.g.,$ mmWave) is a promising direction, as they can provide pose or gait information from different dimensions ($e.g.,$ silhouettes) to better perform recognition tasks.
    To this end, it is crucial to develop unified gait representations that generalize effectively across different data modalities.
    
    \textbf{Cross-Modality Evaluation Protocol.} 
    To enable a fair comparison of person re-ID methods on multi-modal ($e.g.,$ RGB-D) benchmarks, it is imperative to formulate a cross-modality evaluation protocol that standardizes re-ID settings ($e.g.,$ probe/gallery settings, single/multi-shot recognition) of skeleton-based, depth-based, radar-based and other methods.

    \textbf{Skeleton Foundation Model.} LLMs and pose generative models \cite{lucas2022posegpt,bao2024omnipotent} can be synergized to build a skeleton foundation model by training it on large-scale skeletons. It can hopefully be fine-tuned for diverse tasks such as skeleton visualization, augmentation, prediction, etc.     
    Such foundation model will help investigate the limit of adaptability, generality, and interpretability of 3D skeletons, and can potentially serve as a healthcare copilot \cite{zhao2025smart} with expert rules and evidence-based reasoning \cite{zhao2025medrag,zhao2021brain} to empower proactive health monitoring.

\begin{table}[t]
\centering
\scalebox{0.6}{
\renewcommand\arraystretch{1.4}{
% 1.88
\setlength{\tabcolsep}{11.2mm}{
\begin{tabular}{llrr}
\hline
\textbf{Types} & \textbf{Methods} & \textbf{mAP} & \textbf{R$_{1}$} \\ \hline
\multirow{2}{*}{\textbf{Hand-Crafted}} & ${D_{13}}$ & 35.0 & 21.9 \\
 & ${D_{16}}$ & 35.5 & 28.5 \\ \hline
\multirow{3}{*}{\textbf{Sequence-Based}} & PoseGait  & 46.2 & 37.5 \\
 & SimMC  & 45.9 & 60.2 \\
 & Hi-MPC & 47.5 & 64.1 \\ \hline
\multirow{2}{*}{\textbf{Graph-Based}} & MoCos & 47.4 & 57.8 \\
 & TranSG & 48.2 & 60.2 \\ \hline
\end{tabular}
}
}
}
\caption{
 Generalization performance of representative SRID methods on the interdisciplinary benchmark 3DGait — A case study.} 
\label{3DGait_result}
\end{table}

    \textbf{Privacy Protection.} 
    A unique advantage of SRID models is that they inherently protect privacy by avoiding the use of explicit appearance data.
    However, the illegal or irresponsible deployment of person re-ID technologies could jeopardize public security, making it crucial to establish SRID-related laws to protect social safety and personal privacy.

\subsection{Interdisciplinary Prospects}
As shown in Fig. \ref{application_overview}, we list several promising applications of SRID in terms of \textbf{healthcare}, \textbf{embodied AI}, and \textbf{security}:

\label{sec_applications}
\textbf{Healthcare.} Since physiological and psychological states are intrinsically correlated with walking patterns ($e.g.,$ Parkinsonian gait) \cite{lu2025understanding,rao2025survey}, the pre-trained SRID models can be transferred for neurodegenerative disease diagnosis and affective/psychological detection ($e.g.,$ depression) \cite{lu2023see,lu2023epic}. 
    Moreover, they can hopefully support medical gait analysis and rehabilitation assessment ($e.g.,$ stroke) via non-intrusive monitoring in daily environments.

To demonstrate the above interdisciplinary applicability, we follow \cite{rao2025llm} to systematically evaluate various SRID models on a representative healthcare task: Neurodegenerative disease prediction.
The case study in Table \ref{3DGait_result} shows that these models can capture generalizable discriminative gait patterns for this task, suggesting their promising potential for broader real-world healthcare applications.

\textbf{Embodied AI.} 
    With smaller input data and lower resource requirement than RGB-based models, SRID can serve as a fundamental semantic link for intelligent agents and their interaction systems: Agents can act via skeleton-based avatar motion retargeting, perceive through human-centric intention prediction, interact using skeletal gesture based control, and evolve within skeleton-driven synthetic environments, which facilitate robust identity-aware collaboration in robotics.

\textbf{Security.} Leveraging the robustness of skeletal features against appearance variations \cite{han2017space}, SRID advances public safety through privacy-preserving authentication and cross-modal ($e.g.,$ face, fingerprint) identity verification. It also enables forensic behavioral analysis to track suspects and detect abnormal events across varying scenarios particularly where visual details are lacking.

\textbf{Sports Science}. By capturing fine-grained skeletal dynamics, SRID enables biomechanical analysis for athletic performance optimization ($e.g.$, running gait correction and swimming stroke refinement \cite{zhou2026comprehensive}) and sports injury prevention through abnormal movement pattern detection. Furthermore, it provides real-time skill acquisition feedback for interactive coaching ($e.g.$, pose comparison with expert templates), thereby supporting data-driven training in both competitive and recreational sports.

\section{Conclusion}
\label{sec_conclude}
In this paper, we provide the first comprehensive review of 3D skeleton based person re-ID (SRID).
We first define the SRID task and present a timeline to overview the major advancements of SRID. 
Then we formulate a taxonomy of SRID approaches in terms of different skeleton modeling and learning paradigms, systematically presenting foundational mechanisms and reviewing representative approaches. 
An empirical evaluation of state-of-the-art SRID methods is conducted across various benchmarks and evaluation protocols to compare their effectiveness, efficiency, and key properties. 
We further discuss critical challenges along with potential directions, and highlight interdisciplinary prospects of SRID.

\section*{Acknowledgments}
This research is supported by the RIE2025 Industry Alignment Fund – Industry Collaboration Projects (IAF-ICP) (Award I2301E0026), administered by A*STAR, as well as supported by Alibaba Group and NTU Singapore through Alibaba-NTU Global e-Sustainability CorpLab (ANGEL). 
This research is also partially supported by the National Research Foundation, Singapore under its AI Singapore Programme (AISG Award No: AISG2-PhD/2022-01-034[T]).

 % the Joint NTU-UBC Research Centre of Excellence in Active Living for the Elderly (LILY), and

\small{
\bibliographystyle{named}
\bibliography{main}

@String(CVPR= {IEEE Conf. Comput. Vis. Pattern Recog.})

@String(ICCV= {Int. Conf. Comput. Vis.})

@String(ECCV= {Eur. Conf. Comput. Vis.})

@String(ICPR = {Int. Conf. Pattern Recog.})

@String(IJCAI = {IJCAI})

@String(AAAI = {AAAI})

@String(CVPR  = {CVPR})

@String(ICCV  = {ICCV})

@String(ECCV  = {ECCV})

@String(ICPR  = {ICPR})

@article{murray1964walking,
  title={Walking patterns of normal men},
  author={Murray, M Pat and Drought, A Bernard and Kory, Ross C},
  journal={Journal of Bone and Joint Surgery},
  volume={46},
  number={2},
  pages={335--360},
  year={1964},
  publisher={LWW}
}

@inproceedings{haque2016recurrent,
  title={Recurrent attention models for depth-based person identification},
  author={Haque, Albert and Alahi, Alexandre and Fei-Fei, Li},
  booktitle={CVPR},
  pages={1229--1238},
  year={2016}
}

@inproceedings{andersson2015person,
  title={Person identification using anthropometric and gait data from {Kinect} sensor},
  author={Andersson, Virginia O and Araujo, Ricardo M},
  booktitle={Proceedings of the AAAI Conference on Artificial Intelligence (AAAI)},
  pages={425--431},
  year={2015}
}

@inproceedings{munaro20143d,
  title={{3D} reconstruction of freely moving persons for re-identification with a depth sensor},
  author={Munaro, Matteo and Basso, Alberto and Fossati, Andrea and Van Gool, Luc and Menegatti, Emanuele},
  booktitle={International Conference on Robotics and Automation (ICRA)},
  pages={4512--4519},
  year={2014},
  organization={IEEE}
}

@inproceedings{munaro2014feature,
  title={A feature-based approach to people re-identification using skeleton keypoints},
  author={Munaro, Matteo and Ghidoni, Stefano and Dizmen, Deniz Tartaro and Menegatti, Emanuele},
  booktitle={International Conference on Robotics and Automation (ICRA)},
  pages={5644--5651},
  year={2014},
  organization={IEEE}
}

@incollection{munaro2014one,
  title={One-shot person re-identification with a consumer depth camera},
  author={Munaro, Matteo and Fossati, Andrea and Basso, Alberto and Menegatti, Emanuele and Van Gool, Luc},
  booktitle={Person Re-Identification},
  pages={161--181},
  year={2014},
  publisher={Springer}
}

@article{pala2019enhanced,
  title={Enhanced skeleton and face {3D} data for person re-identification from depth cameras},
  author={Pala, Pietro and Seidenari, Lorenzo and Berretti, Stefano and Del Bimbo, Alberto},
  journal={Computers \& Graphics},
  volume={79},
  pages={69--80},
  year={2019},
  publisher={Elsevier}
}

@article{han2017space,
  title={Space-time representation of people based on {3D} skeletal data: A review},
  author={Han, Fei and Reily, Brian and Hoff, William and Zhang, Hao},
  journal={Computer Vision and Image Understanding},
  volume={158},
  pages={85--105},
  year={2017},
  publisher={Elsevier}
}

@article{nambiar2019gait,
  title={Gait-based Person Re-identification: A Survey},
  author={Nambiar, Athira and Bernardino, Alexandre and Nascimento, Jacinto C},
  journal={ACM Computing Surveys},
  volume={52},
  number={2},
  pages={33},
  year={2019},
  publisher={ACM}
}

@article{kastaniotis2015framework,
  title={A framework for gait-based recognition using {Kinect}},
  author={Kastaniotis, Dimitris and Theodorakopoulos, Ilias and Theoharatos, Christos and Economou, George and Fotopoulos, Spiros},
  journal={Pattern Recognition Letters},
  volume={68},
  pages={327--335},
  year={2015},
  publisher={Elsevier}
}

@article{liao2020model,
  title={A model-based gait recognition method with body pose and human prior knowledge},
  author={Liao, Rijun and Yu, Shiqi and An, Weizhi and Huang, Yongzhen},
  journal={Pattern Recognition},
  volume={98},
  pages={107069},
  year={2020},
  publisher={Elsevier}
}

@inproceedings{barbosa2012re,
  title={Re-identification with {RGB-D} sensors},
  author={Barbosa, Igor Barros and Cristani, Marco and Del Bue, Alessio and Bazzani, Loris and Murino, Vittorio},
  booktitle={ECCV Workshop},
  pages={433--442},
  year={2012},
  organization={Springer}
}

@inproceedings{yoo2002extracting,
  title={Extracting gait signatures based on anatomical knowledge},
  author={Yoo, Jang-Hee and Nixon, Mark S and Harris, Chris J},
  booktitle={Proceedings of BMVA Symposium on Advancing Biometric Technologies},
  pages={596--606},
  year={2002},
  organization={Citeseer}
}

@inproceedings{gharghabi2015people,
  title={People re-identification using {3D} descriptor with skeleton information},
  author={Gharghabi, Shaghayegh and Shamshirdar, Faraz and Shangari, \textit{et al}, Taher Abbas},
  booktitle={2015 International Conference on Informatics, Electronics \& Vision (ICIEV)},
  pages={1--5},
  year={2015},
  organization={IEEE}
}

@inproceedings{nambiar2017context,
  title={Context-aware person re-identification in the wild via fusion of gait and anthropometric features},
  author={Nambiar, Athira and Bernardino, Alexandre and Nascimento, Jacinto C and Fred, Ana},
  booktitle={International Conference on Automatic Face \& Gesture Recognition},
  pages={973--980},
  year={2017},
  organization={IEEE}
}

@inproceedings{shotton2011real-time,
title={Real-time human pose recognition in parts from single depth images},
author={Shotton, Jamie and Fitzgibbon, Andrew and Cook, Mat and Sharp, Toby and Finocchio, Mark J and Moore, Richard and Kipman, Alex Abenathar and Blake, Andrew},
booktitle={CVPR},
pages={1297--1304},
year={2011}}

@inproceedings{rao2020self,
	title="Self-Supervised Gait Encoding with Locality-Aware Attention for Person Re-Identification",
	author="Haocong {Rao} and Siqi {Wang} and Xiping {Hu} and Mingkui {Tan} and Huang {Da} and Jun {Cheng} and Bin {Hu}",
	booktitle="International Joint Conference on Artificial Intelligence (IJCAI)",
	volume="1",
	pages="898--905",
	year="2020"
}

@inproceedings{li2020dynamic,
	title="Dynamic Multiscale Graph Neural Networks for {3D} Skeleton Based Human Motion Prediction",
	author="Maosen {Li} and Siheng {Chen} and Yangheng {Zhao} and Ya {Zhang} and Yanfeng {Wang} and Qi {Tian}",
	booktitle="CVPR",
	pages="214--223",
	year="2020"
}

@inproceedings{yan2018spatial,
	title="Spatial Temporal Graph Convolutional Networks for Skeleton-Based Action Recognition.",
	author="Sijie {Yan} and Yuanjun {Xiong} and Dahua {Lin}",
	booktitle="Proceedings of the AAAI Conference on Artificial Intelligence (AAAI)",
	pages="7444--7452",
	year="2018"
}

@article{rao2021self,
  title={A self-supervised gait encoding approach with locality-awareness for {3D} skeleton based person re-identification},
  author={Rao, Haocong and Wang, Siqi and Hu, Xiping and Tan, Mingkui and Guo, Yi and Cheng, Jun and Liu, Xinwang and Hu, Bin},
  journal={IEEE Transactions on Pattern Analysis and Machine Intelligence},
  volume={44},
  number={10},
  pages={6649--6666},
  year={2022},
  publisher={IEEE}
}

@article{cao2019openpose,
  title={{OpenPose}: Realtime multi-person {2D} pose estimation using Part Affinity Fields},
  author={Cao, Zhe and Hidalgo, Gines and Simon, Tomas and Wei, Shih-En and Sheikh, Yaser},
  journal={IEEE Transactions on Pattern Analysis and Machine Intelligence},
  volume={43},
  number={1},
  pages={172--186},
  year={2019},
  publisher={IEEE}
}

@inproceedings{chen20173d,
  title={{3D} human pose estimation= {2D} pose estimation+ matching},
  author={Chen, Ching-Hang and Ramanan, Deva},
  booktitle={CVPR},
  pages={7035--7043},
  year={2017}
}

@inproceedings{rao2021multi,
  title={Multi-Level Graph Encoding with Structural-Collaborative Relation Learning for Skeleton-Based Person Re-Identification},
  author={Rao, Haocong and Xu, Shihao and Hu, Xiping and Cheng, Jun and Hu, Bin},
  booktitle={International Joint Conference on Artificial Intelligence (IJCAI)},
  pages={973--980},
  year={2021}
}

@inproceedings{rao2021sm,
  title={{SM-SGE}: A Self-Supervised Multi-Scale Skeleton Graph Encoding Framework for Person Re-Identification},
  author={Rao, Haocong and Hu, Xiping and Cheng, Jun and Hu, Bin},
  booktitle={Proceedings of the 29th ACM International Conference on Multimedia},
  pages={1812--1820},
  year={2021}
}

@inproceedings{zheng2015scalable,
  title={Scalable person re-identification: A benchmark},
  author={Zheng, Liang and Shen, Liyue and Tian, Lu and Wang, Shengjin and Wang, Jingdong and Tian, Qi},
  booktitle={ICCV},
  pages={1116--1124},
  year={2015}
}

@inproceedings{farenzena2010person,
  title={Person re-identification by symmetry-driven accumulation of local features},
  author={Farenzena, Michela and Bazzani, Loris and Perina, Alessandro and Murino, Vittorio and Cristani, Marco},
  booktitle={CVPR},
  pages={2360--2367},
  year={2010},
  organization={IEEE}
}

@article{ye2021deep,
  title={Deep learning for person re-identification: A survey and outlook},
  author={Ye, Mang and Shen, Jianbing and Lin, Gaojie and Xiang, Tao and Shao, Ling and Hoi, Steven CH},
  journal={IEEE Transactions on Pattern Analysis and Machine Intelligence},
  volume={44},
  number={6},
  pages={2872--2893},
  year={2021},
  publisher={IEEE}
}

@article{hermans2017defense,
  title={In defense of the triplet loss for person re-identification},
  author={Hermans, Alexander and Beyer, Lucas and Leibe, Bastian},
  journal={arXiv preprint arXiv:1703.07737},
  year={2017}
}

@inproceedings{rao2022simmc,
  title={{SimMC}: Simple Masked Contrastive Learning of Skeleton Representations for Unsupervised Person Re-Identification},
  author={Rao, Haocong and Miao, Chunyan},
  booktitle={International Joint Conference on Artificial Intelligence (IJCAI)},
  pages={1290--1297},
  year={2022}
}

@article{rao2022skeleton,
  title={Skeleton Prototype Contrastive Learning with Multi-Level Graph Relation Modeling for Unsupervised Person Re-Identification},
  author={{R}ao, Haocong and Miao, Chunyan},
  journal={arXiv preprint arXiv:2208.11814},
  year={2022}
}

@inproceedings{rao2023transg,
  title={{TranSG}: Transformer-Based Skeleton Graph Prototype Contrastive Learning with Structure-Trajectory Prompted Reconstruction for Person Re-Identification},
  author={{Rao}, Haocong and Miao, Chunyan},
  booktitle={CVPR},
  year={2023}
}

@article{rao2024hierarchical,
  title={Hierarchical skeleton meta-prototype contrastive learning with hard skeleton mining for unsupervised person re-identification},
  author={Rao, Haocong and Leung, Cyril and Miao, Chunyan},
  journal={International Journal of Computer Vision},
  volume={132},
  number={1},
  pages={238--260},
  year={2024},
  publisher={Springer}
}

@inproceedings{bondi2018long,
  title={Long term person re-identification from depth cameras using facial and skeleton data},
  author={Bondi, Enrico and Pala, Pietro and Seidenari, Lorenzo and Berretti, Stefano and Del Bimbo, Alberto},
  booktitle={International Conference on Pattern Recognition (ICPR) Workshop},
  pages={29--41},
  year={2016},
}

@article{elaoud2021person,
  title={Person Re-Identification from different views based on dynamic linear combination of distances},
  author={Elaoud, Amani and Barhoumi, Walid and Drira, Hassen and Zagrouba, Ezzeddine},
  journal={Multimedia Tools and Applications},
  volume={80},
  pages={17685--17704},
  year={2021},
  publisher={Springer}
}

@inproceedings{chen2022keypoint,
  title={Keypoint message passing for video-based person re-identification},
  author={Chen, Di and D{\"o}ring, Andreas and Zhang, Shanshan and Yang, Jian and Gall, Juergen and Schiele, Bernt},
  booktitle={Proceedings of the AAAI Conference on Artificial Intelligence (AAAI)},
  volume={36},
  number={1},
  pages={239--247},
  year={2022}
}

@article{pala2015multimodal,
  title={Multimodal person reidentification using {RGB-D} cameras},
  author={Pala, Federico and Satta, Riccardo and Fumera, Giorgio and Roli, Fabio},
  journal={IEEE Transactions on Circuits and Systems for Video Technology},
  volume={26},
  number={4},
  pages={788--799},
  year={2015},
  publisher={IEEE}
}

@article{rashmi2022human,
  title={Human identification system using {3D} skeleton-based gait features and {LSTM} model},
  author={Rashmi, M and Guddeti, Ram Mohana Reddy},
  journal={Journal of Visual Communication and Image Representation (JVCIR)},
  volume={82},
  pages={103416},
  year={2022},
  publisher={Elsevier}
}

@inproceedings{wei2020person,
  title={Person Identification by Walking Gesture Using Skeleton Sequences},
  author={Wei, Chu-Chien and Tsai, Li-Huang and Chou, Hsin-Ping and Chang, Shih-Chieh},
  booktitle={Advanced Concepts for Intelligent Vision Systems},
  pages={205--214},
  year={2020},
  organization={Springer}
}

@article{huynh2020learning,
  title={Learning {3D} spatiotemporal gait feature by convolutional network for person identification},
  author={Huynh-The, Thien and Hua, Cam-Hao and Tu, Nguyen Anh and Kim, Dong-Seong},
  journal={Neurocomputing},
  volume={397},
  pages={192--202},
  year={2020},
  publisher={Elsevier}
}

@article{kastaniotis2016gait,
  title={Gait based recognition via fusing information from Euclidean and Riemannian manifolds},
  author={Kastaniotis, Dimitris and Theodorakopoulos, Ilias and Economou, George and Fotopoulos, Spiros},
  journal={Pattern Recognition Letters},
  volume={84},
  pages={245--251},
  year={2016},
  publisher={Elsevier}
}

@inproceedings{wei2018person,
  title={Person transfer gan to bridge domain gap for person re-identification},
  author={Wei, Longhui and Zhang, Shiliang and Gao, Wen and Tian, Qi},
  booktitle={CVPR},
  pages={79--88},
  year={2018}
}

@article{ho2020denoising,
  title={Denoising diffusion probabilistic models},
  author={Ho, Jonathan and Jain, Ajay and Abbeel, Pieter},
  journal={Advances in Neural Information Processing Systems (NeurIPS)},
  volume={33},
  pages={6840--6851},
  year={2020}
}

@inproceedings{yan2017skeleton,
  title={Skeleton-aided articulated motion generation},
  author={Yan, Yichao and Xu, Jingwei and Ni, Bingbing and Zhang, Wendong and Yang, Xiaokang},
  booktitle={Proceedings of the 25th ACM international conference on Multimedia},
  pages={199--207},
  year={2017}
}

@inproceedings{zhou2016learning,
  title={Learning deep features for discriminative localization},
  author={Zhou, Bolei and Khosla, Aditya and Lapedriza, Agata and Oliva, Aude and Torralba, Antonio},
  booktitle={CVPR},
  pages={2921--2929},
  year={2016}
}

@article{ji2021survey,
  title={A survey on knowledge graphs: Representation, acquisition, and applications},
  author={Ji, Shaoxiong and Pan, Shirui and Cambria, Erik and Marttinen, Pekka and Philip, S Yu},
  journal={IEEE Transactions on Neural Networks and Learning Systems},
  volume={33},
  number={2},
  pages={494--514},
  year={2021},
  publisher={IEEE}
}

@inproceedings{lucas2022posegpt,
  title={{PoseGPT}: Quantization-based {3D} human motion generation and forecasting},
  author={Lucas, Thomas and Baradel, Fabien and Weinzaepfel, Philippe and Rogez, Gr{\'e}gory},
  booktitle={ECCV},
  pages={417--435},
  year={2022},
  organization={Springer}
}

@inproceedings{rao2025motif,
  title={Motif Guided Graph Transformer with Combinatorial Skeleton Prototype Learning for Skeleton-Based Person Re-Identification},
  author={Rao, Haocong and Miao, Chunyan},
  booktitle={Proceedings of the AAAI Conference on Artificial Intelligence (AAAI)},
  year={2025}
}

@article{rao2025survey,
  title={A Survey of Artificial Intelligence in Gait-Based Neurodegenerative Disease Diagnosis},
  author={Rao, Haocong and Zeng, Minlin and Zhao, Xuejiao and Miao, Chunyan},
  journal={Neurocomputing},
  year={2025}
}

@article{cunado2003automatic,
  title={Automatic extraction and description of human gait models for recognition purposes},
  author={Cunado, David and Nixon, Mark S and Carter, John N},
  journal={Computer vision and image understanding},
  volume={90},
  number={1},
  pages={1--41},
  year={2003},
  publisher={Elsevier}
}

@inproceedings{he2022masked,
  title={Masked autoencoders are scalable vision learners},
  author={He, Kaiming and Chen, Xinlei and Xie, Saining and Li, Yanghao and Doll{\'a}r, Piotr and Girshick, Ross},
  booktitle={CVPR},
  pages={16000--16009},
  year={2022}
}

@inproceedings{elaoud2017analysis,
  title={Analysis of skeletal shape trajectories for person re-identification},
  author={Elaoud, Amani and Barhoumi, Walid and Drira, Hassen and Zagrouba, Ezzeddine},
  booktitle={International Conference on Advanced Concepts for Intelligent Vision Systems},
  pages={138--149},
  year={2017},
  organization={Springer}
}

@article{zhang2023spatial,
  title={Spatial transformer network on skeleton-based gait recognition},
  author={Zhang, Cun and Chen, Xing-Peng and Han, Guo-Qiang and Liu, Xiang-Jie},
  journal={Expert Systems},
  volume={40},
  number={6},
  pages={e13244},
  year={2023},
  publisher={Wiley Online Library}
}

@inproceedings{chen2021channel,
  title={Channel-wise topology refinement graph convolution for skeleton-based action recognition},
  author={Chen, Yuxin and Zhang, Ziqi and Yuan, Chunfeng and Li, Bing and Deng, Ying and Hu, Weiming},
  booktitle={ICCV},
  pages={13359--13368},
  year={2021}
}

@inproceedings{fu2023gpgait,
  title={{GPGait}: Generalized pose-based gait recognition},
  author={Fu, Yang and Meng, Shibei and Hou, Saihui and Hu, Xuecai and Huang, Yongzhen},
  booktitle={ICCV},
  pages={19595--19604},
  year={2023}
}

@inproceedings{fan2024skeletongait,
  title={{SkeletonGait}: Gait recognition using skeleton maps},
  author={Fan, Chao and Ma, Jingzhe and Jin, Dongyang and Shen, Chuanfu and Yu, Shiqi},
  booktitle={Proceedings of the AAAI Conference on Artificial Intelligence (AAAI)},
  volume={38},
  number={2},
  pages={1662--1669},
  year={2024}
}

@inproceedings{rao2025llm,
  title={LLM-Powered Interpretable 3D Gait Visualization and Analysis Platform for Interdisciplinary AI Applications},
  author={Rao, Haocong and Zhao, Jiachen and Miao, Chunyan},
  booktitle={27th International Conference on Human-Computer Interaction (HCII)},
  year={2025}
}

@inproceedings{wang2023amai,
  title={Video-based gait analysis for assessing Alzheimer’s Disease and Dementia with Lewy Bodies},
  author={Wang, Diwei and Zouaoui, Chaima and Jang, Jinhyeok and Drira, Hassen and Seo, Hyewon},
  booktitle={MICCAI Workshop on Applications of Medical AI},
  pages={72--82},
  year={2023},
  organization={Springer}
}

@inproceedings{lu2025understanding,
  title={Understanding emotional body expressions via large language models},
  author={Lu, Haifeng and Chen, Jiuyi and Liang, Feng and Tan, Mingkui and Zeng, Runhao and Hu, Xiping},
  booktitle={Proceedings
of the AAAI Conference on Artificial Intelligence (AAAI)},
  volume={39},
  number={2},
  pages={1447--1455},
  year={2025}
}

@inproceedings{lu2023see,
  title={See your emotion from gait using unlabeled skeleton data},
  author={Lu, Haifeng and Hu, Xiping and Hu, Bin},
  booktitle={Proceedings
of the AAAI Conference on Artificial Intelligence (AAAI)},
  volume={37},
  number={2},
  pages={1826--1834},
  year={2023}
}

@article{lu2023epic,
  title={EPIC: Emotion perception by spatio-temporal interaction context of gait},
  author={Lu, Haifeng and Xu, Shihao and Zhao, Shipeng and Hu, Xiping and Ma, Rong and Hu, Bin},
  journal={IEEE Journal of Biomedical and Health Informatics},
  volume={28},
  number={5},
  pages={2592--2601},
  year={2023},
  publisher={IEEE}
}

@inproceedings{zhao2025medrag,
  title={Medrag: Enhancing retrieval-augmented generation with knowledge graph-elicited reasoning for healthcare copilot},
  author={Zhao, Xuejiao and Liu, Siyan and Yang, Su-Yin and Miao, Chunyan},
  booktitle={Proceedings of the ACM on Web Conference 2025},
  pages={4442--4457},
  year={2025}
}

@article{zhao2025smart,
  title={A smart multimodal healthcare copilot with powerful {LLM} reasoning},
  author={Zhao, Xuejiao and Liu, Siyan and Yang, Su-Yin and Miao, Chunyan},
  journal={arXiv preprint arXiv:2506.02470},
  year={2025}
}

@article{zhao2021brain,
  title={Brain-inspired search engine assistant based on knowledge graph},
  author={Zhao, Xuejiao and Chen, Huanhuan and Xing, Zhenchang and Miao, Chunyan},
  journal={IEEE Transactions on Neural Networks and Learning Systems},
  volume={34},
  number={8},
  pages={4386--4400},
  year={2021},
  publisher={IEEE}
}

@inproceedings{bao2026activityforensics,
title={ActivityForensics: A Comprehensive Benchmark for Localizing Manipulated Activity in Videos},
author={Bao, Peijun and Luo, Anwei and Pan, Gang and Kot, Alex C. and Jiang, Xudong},
booktitle={Proceedings of the IEEE/CVF Conference on Computer Vision and Pattern Recognition (CVPR)},
year={2026}
}

@inproceedings{bao2025vid,
title={ Vid-Group: Temporal Video Grounding Pretraining from Unlabeled Videos in the Wild},
author={ Peijun Bao and Chenqi Kong and Siyuan Yang and Zihao Shao and Xinghao Jiang and Boon Poh Ng and Meng Hwa Er and Alex C. Kot },
booktitle={2025 IEEE/CVF International Conference on Computer Vision (ICCV)},
year={2025}
}

@inproceedings{bao2024omnipotent,
title={Omnipotent Distillation with LLMs for Weakly-Supervised Natural Language Video Localization: When Divergence Meets Consistency},
author={Bao, Peijun and Shao, Zihao and Yang, Wenhan and Ng, Boon Poh and Er, Meng Hwa and Kot, Alex C},
booktitle={Proceedings of the AAAI Conference on Artificial Intelligence (AAAI)},
year={2024}
}

@article{zhou2026comprehensive,
author = {Kanglei Zhou and Ruizhi Cai and Liyuan Wang and
Hubert P. H. Shum and Xiaohui Liang},
title = {A comprehensive survey of action quality assessment:
Method and benchmark},
journal = {Pattern Recognition},
year = {2026},
pages = {113933},
doi = {10.1016/j.patcog.2026.113933},
note = {Review article}
}
}

\end{document}